\documentclass[10pt,twocolumn,letterpaper]{article}

\usepackage{iccv}
\usepackage{times}
\usepackage{epsfig}
\usepackage{graphicx}
\usepackage{amsmath}
\usepackage{amssymb}
\usepackage{multirow}
\usepackage{xcolor}
\definecolor{b-lstm}{RGB}{244,15,72}
\definecolor{ours}{RGB}{16,80,242}
\definecolor{pie}{RGB}{242,200,16}
\definecolor{gt}{RGB}{16,244,15}
\definecolor{no_cim}{RGB}{239,7,206}
\definecolor{hybrid}{RGB}{87,225,239}
\graphicspath{{images/}}

\usepackage[breaklinks=true,bookmarks=false]{hyperref}

\iccvfinalcopy 

\def\iccvPaperID{4350} 

\ificcvfinal\fi

\begin{document}

\title{Bifold and Semantic Reasoning for Pedestrian Behavior Prediction}

\author{Amir Rasouli, Mohsen Rohani, Jun Luo\\
Huawei Technologies Canada\\
{\tt\small \{amir.rasouli, mohsen.rohani, jun.luo1\}@huawei.com}}

\maketitle
\ificcvfinal\thispagestyle{empty}\fi

\begin{abstract}
Pedestrian behavior prediction is one of the major challenges for intelligent driving systems. Pedestrians often exhibit complex behaviors influenced by various contextual elements. To address this problem, we propose BiPed, a multitask learning framework that simultaneously predicts trajectories and actions of pedestrians by relying on multimodal data. Our method benefits from 1) a bifold encoding approach where different data modalities are processed independently allowing them to develop their own representations, and jointly to produce a representation for all modalities using shared parameters; 2) a novel interaction modeling technique that relies on categorical semantic parsing of the scenes to capture interactions between target pedestrians and their surroundings; and 3) a bifold prediction mechanism that uses both independent and shared decoding of multimodal representations. Using public pedestrian behavior benchmark datasets for driving, PIE and JAAD, we highlight the benefits of the proposed method for behavior prediction and show that our model achieves state-of-the-art performance and improves trajectory and action prediction by up to $22\%$ and $9\%$ respectively. We further investigate the contributions of the proposed reasoning techniques via extensive ablation studies.
\end{abstract}
\vspace{-0.5cm}
\section{Introduction}
Predicting road user behavior in complex urban environments is fundamental for assistive and intelligent driving systems. Prediction is particularly challenging when these systems are encountering pedestrians who exhibit diverse behaviors \cite{Rasouli_2017_IV} that depend on various contextual factors, such as social interactions, road structure, traffic condition, and other environmental factors \cite{Rasouli_2019_ITS}. 

\begin{figure}
\centering
\includegraphics[width=0.83\columnwidth]{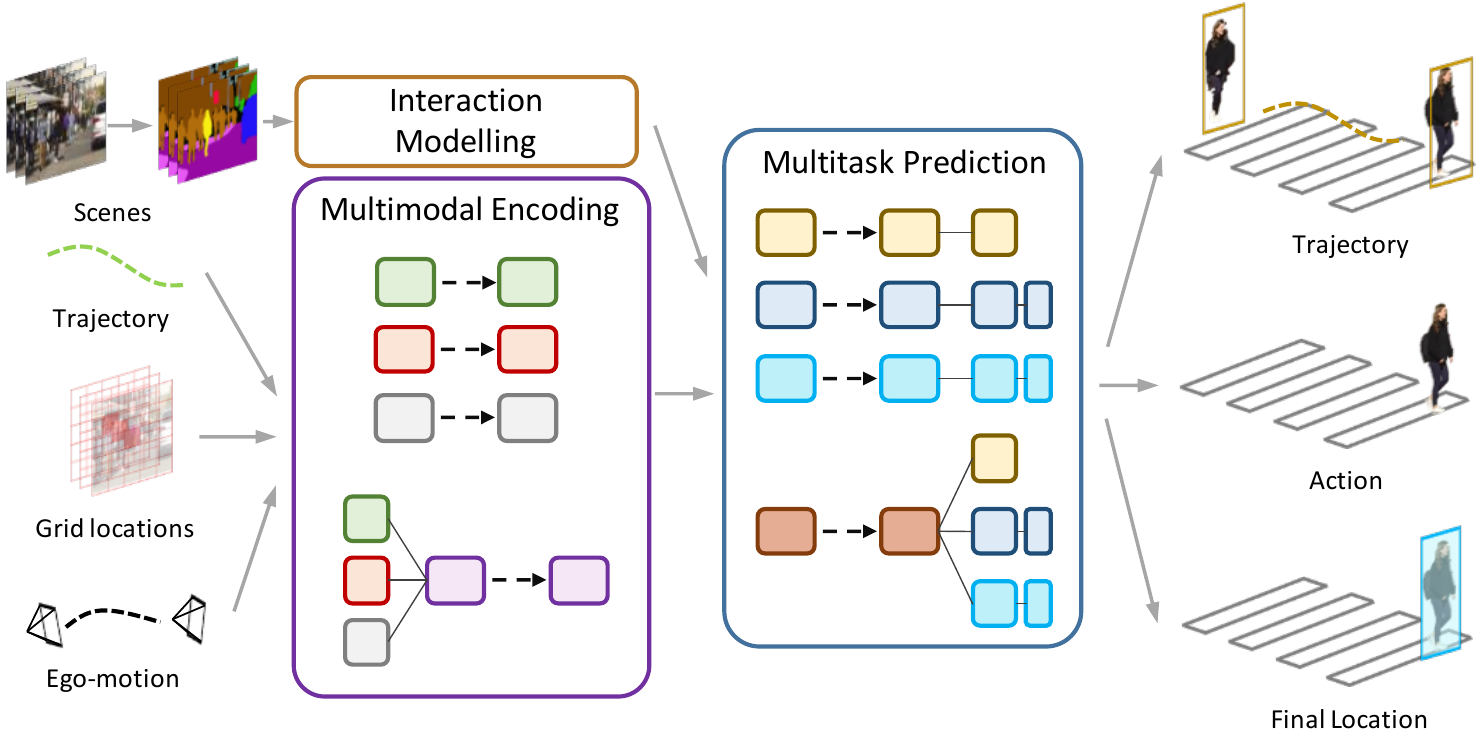}
\caption{Proposed multitask learning for simultaneous prediction of pedestrian trajectory, action and final location. Interaction of traffic elements are modeled along with separately and jointly encoded visual context, pedestrian motion, and ego-motion inputs.}
\vspace{-0.5cm}
\label{first_image}
\end{figure}

Pedestrian behavior can be predicted implicitly in the form of future trajectories \cite{Sun_2020_CVPR,Makansi_2020_CVPR,Malla_2020_CVPR}, or explicitly in the form of upcoming actions \cite{Chaabane_2020_WACV,Liu_2020_RAL,Rasouli_2019_BMVC}. It is evident from recent studies \cite{Malla_2020_CVPR,Liang_2019_CVPR,Rasouli_2019_ICCV} that both types of behavior prediction play complementary roles. For instance, predicting pedestrian actions, such as crossing the road, implies the possibility of a lateral motion across the road. Similarly, a pedestrian approaching a parked vehicle is expected to interact with it. To capture these complementary aspects of behavior, we propose a multitask learning framework that simultaneously predicts trajectories, actions, and final locations of pedestrians (see Figure \ref{first_image}). To learn complex pedestrian behavior, our model relies on multiple data modalities including visual context, pedestrian motion, and ego-vehicle dynamics. The proposed method independently and jointly processes different input modalities and tasks. Independent processing allows each modality or task to learn its own  parameters, whereas joint processing acts as a regularizer, inducing the model to learn more representative features. Since pedestrians' behaviors are often influenced by what is around them, we introduce a novel technique to model interactions between target pedestrians and their surroundings based on the semantic composition of the scenes. The proposed technique relies on visuospatial semantic representations of the scenes divided into categories based on object classes. 


We evaluate the performance of the proposed method using public pedestrian behavior benchmark datasets, PIE \cite{Rasouli_2019_ICCV} and JAAD \cite{Rasouli_2017_ICCVW}, and show that our method significantly improves over state-of-the-art algorithms on both pedestrian trajectory and action prediction tasks.
\vspace{-0.2cm}
\section{Related Works}
\subsection{Multimodal Behavior Prediction}
Human behavior prediction research in computer vision and robotics has many practical applications, such as human-object \cite{Liu_2020_ECCV,Piergiovanni_2020_ECCV} and human-human \cite{Joo_2019_CVPR,Yao_2018_CVPR} interaction, risk assessment \cite{Strickland_2018_ICRA,Zeng_2017_CVPR}, anomaly detection \cite{Epstein_2020_CVPR}, surveillance \cite{Ma_2020_ECCV,Liang_2019_CVPR}, sports forecasting \cite{Qi_2020_CVPR,Felsen_2017_ICCV}, and intelligent driving systems \cite{Fang_2020_CVPR,Rasouli_2019_BMVC}. As for pedestrians, their behaviors can be predicted implicitly in terms of trajectories \cite{Hu_2020_CVPR,Mohamed_2020_CVPR,Sun_2020_CVPR,Sun_2020_CVPR_2,Mangalam_2020_ECCV} or explicitly in terms of actions, such as crossing the road \cite{Aliakbarian_2018_ACCV,Rasouli_2019_BMVC}, or interacting with objects \cite{Liang_2019_CVPR}.

\textbf{Trajectory Prediction.} 
Pedestrian prediction research is dominated by trajectory prediction. A large body of work in this domain is dedicated to prediction on surveillance sequences where the movements of groups of pedestrians are observed from a fixed bird's eye view perspective  \cite{Hu_2020_CVPR,Mohamed_2020_CVPR,Sun_2020_CVPR,Sun_2020_CVPR_2,Mangalam_2020_ECCV,Choi_2019_ICCV,Zhang_2019_CVPR,Sadeghian_2019_CVPR,Gupta_2018_CVPR}. 

Opposed to bird's eye view prediction algorithms are ego-centric methods that predict the trajectories of pedestrians in the image plane recorded from the perspective of a moving camera \cite{Makansi_2020_CVPR,Malla_2020_CVPR,Rasouli_2019_ICCV,Yagi_2018_CVPR,Yao_2019_ICRA,
Yao_2019_IROS,Bhattacharyya_2018_CVPR,Chandra_2019_CVPR}. Ego-centric trajectory prediction is generally more challenging because, first, in the absence of depth information, the relative positions of agents are difficult to infer and, second, the ego-motion of the camera can influence both the behavior of pedestrians and the predicted trajectories in the image plane. To address these challenges, ego-centric algorithms use multimodal approaches. For example, the model in \cite{Bhattacharyya_2018_CVPR} proposes a two-stream recurrent encoder-decoder architecture where one stream processes pedestrian trajectories and the other ego-motion of the vehicle. The output of the ego-motion predictor in conjunction with the pedestrian stream is used to forecast trajectories. Some models based on a similar architecture use pedestrian action (e.g. waiting to cross \cite{Malla_2020_CVPR}) or intention of performing an action (e.g. crossing the road \cite{Rasouli_2019_ICCV}) as inputs to infer future trajectories. Some methods rely on fully feedforward approaches, such as \cite{Yagi_2018_CVPR}, where three streams of 1D convolutional layers encode ego-motion, pedestrians' trajectories, and poses, the output of which are decoded using a convolutional decoder. 

 
\textbf{Action Prediction.}
Pedestrian action prediction is also being actively investigated with an emphasis on prediction of interactions between humans or groups \cite{Zhao_2019_ICCV,Yao_2018_CVPR,Chen_2018_ECCV,Shi_2018_ECCV,Kong_2017_CVPR,Aliakbarian_2017_ICCV}. In the driving context, the main focus is on predicting pedestrian crossing action to assess the risks and anticipate pedestrian trajectories for safe motion planning  \cite{Kotseruba_2021_WACV,Chaabane_2020_WACV,Saleh_2019_ICRA,Gujjar_2019_ICRA,Rasouli_2017_ICCVW,Liu_2020_RAL,Rasouli_2019_BMVC,Aliakbarian_2018_ACCV}. A group of these algorithms use unimodal image sequences for prediction. For example, the methods of \cite{Gujjar_2019_ICRA,Chaabane_2020_WACV} are convolutional generative models that predict future traffic scenes, which are then used to predict crossing events. In \cite{Saleh_2019_ICRA} a 3D DenseNet is used to first localize pedestrians and then predict their crossing actions. 

For crossing prediction, multimodal architectures are more common \cite{Rasouli_2017_ICCVW,Liu_2020_RAL,Rasouli_2019_BMVC,Aliakbarian_2018_ACCV}. For instance, the model in \cite{Rasouli_2019_BMVC} is a multi-level recurrent architecture, which receives as input the appearance of pedestrians and their surrounding context, pedestrians' trajectories and poses as well as the ego-vehicle speed. The features are gradually fed into the network at each level according to their complexity. The method proposed in \cite{Aliakbarian_2018_ACCV} uses a hierarchical LSTM architecture in which visual features, including optical flow maps and images, and vehicle dynamics are encoded using independent LSTMs. The outputs of these LSTMs are concatenated and fed into an embedding layer followed by another LSTM prior to prediction. The method in \cite{Kotseruba_2021_WACV} is a hybrid architecture that encodes visual features using a 3D convolution network and other modalities using LSTMs followed by temporal attention modules. The representations are fed into a modality attention unit prior to prediction.

Independent encoding of different data modalities, as proposed previously, however, is more susceptible to noise and may not capture cross-modal correlations. Hence, we propose a bifold approach that encodes different modalities both independently and jointy, thus inducing the system to learn more representative features for different tasks, while learning temporal correlations between different modalities. 

\subsection{Interaction Modeling}
One of the fundamental components of behavior prediction in a multi-agent setting is the ability to understand the interactions between agents as their behaviors can potentially impact one another. Thus, interaction modeling is widely used for pedestrian trajectory prediction. For instance, the methods in \cite{Gupta_2018_CVPR,Sun_2020_CVPR_2,Mangalam_2020_ECCV} use the social pooling technique, which jointly processes trajectories of pedestrians within a neighboring region to learn the spatial dependencies among them. Attention-based approaches assign importance values to interacting pedestrians according to their relative distance and motion \cite{Sadeghian_2019_CVPR,Zhang_2019_CVPR,Park_2020_ECCV}. Alternatively, graph structures can be used to assign importance to interacting pedestrians represented as nodes \cite{Li_2020_NeurIPS,Mohamed_2020_CVPR,Sun_2020_CVPR,Kosaraju_2019_NeurIPS,Kipf_2018_ICML}. 

Ego-centric trajectory prediction algorithms follow a similar trajectory-based route to model interactions. For example, the method in \cite{Chandra_2019_CVPR} defines two regions for modeling interactions between traffic participants: a region around the ego-vehicle defined by an ellipsoid on the image plane and a driving horizon of the vehicle that is determined based on the actions of the driver. The authors of \cite{Malla_2020_CVPR} model the interactions by jointly processing the past locations and action information of all agents in the scene using an RNN. 

In the absence of global positions of objects, modeling interactions based on 2D coordinates in the image plane can be problematic. For example, two people of different heights walking next to each other can have  trajectories similar to those of two people of the same height but far apart from each other. In addition, pair-wise modeling of interactions between each pedestrian and all other road users is not scalable, especially in urban driving scenarios where many traffic participants potentially interact with each other. To address these shortcomings, we propose a categorical interaction modeling technique, which relies on the visuospatial changes of different categories of objects over time. Our method encodes both dynamic and static context of the scenes and uses an attention mechanism to determine the importance of different contextual elements.

\subsection{Multitask Learning}
As evidenced by various domains of machine learning, such as action/expression understanding \cite{Luvizon_2018_CVPR,Guo_2018_ECCV,Hu_2018_ECCV,Du_2019_CVPR}, object recognition \cite{Mallya_2018_ECCV,Zeng_2019_ICCV,Tang_2019_ICCV,Hassani_2019_ICCV}, intelligent driving \cite{Casas_2018_CORL,Kendall_2018_CVPR,Liang_2019_CVPR_2,Liu_2019_CVPR,Wu_2020_CVPR_2}, and other computer vision applications \cite{Xu_2018_CVPR,Zhao_2018_ECCV,Gao_2019_CVPR,Bragman_2019_ICCV}, multitask learning is an effective way for improving the performance on multiple tasks. 

Pedestrian behavior prediction is not an exception and in recent years a number of multitask algorithms have been proposed to solve this problem \cite{Hasan_2018_CVPR,Fernando_2018_ACCV,Liang_2019_CVPR,Zhang_2020_CVPR}. For example, the authors of \cite{Hasan_2018_CVPR} simultaneously predict pedestrian head poses and trajectories and exploit the correlation between the two to improve trajectory prediction. The method in \cite{Zhang_2020_CVPR} detects pedestrians and predicts their future trajectories at the same time . It uses point cloud sequences encoded into different feature representations which are fed into a feedforward backbone network. The output of the backbone is used to generate temporal proposals for localizing pedestrians and predicting their trajectories. Liang et al. \cite{Liang_2019_CVPR}, jointly predict pedestrian trajectory and activity, e.g. interacting with a car, using a recurrent framework that encodes different data types, including scene semantics, poses, and trajectories, and combines their representations for joined reasoning on different tasks in separate prediction branches.

In the context of intelligent driving, predicting both trajectories and actions of pedestrians is important for planning. These tasks can play complementary roles -- trajectory prediction provides accurate location information in future frames, whereas action prediction helps interpret the nature of events and the types of motions to be expected. Given such a mutually beneficial relationship between these two tasks, we propose a multitask learning approach that predicts trajectories and actions simultaneously. Unlike other approaches, in addition to independent task predictors, we use a shared prediction module to capture correlation across different tasks, and make final predictions based on the outputs of both independent and joint modules.

\noindent\textbf{Contributions:} 1) We propose a multitask pedestrian behavior prediction framework to simultaneously predict pedestrian trajectory, action and final location in urban traffic scenes using ego-centric image sequences. The proposed approach benefits from a bifold  encoding scheme that independently and jointly learns different data modalities, a categorical interaction module which encodes the interactions between target pedestrians and their surroundings, and a prediction mechanism which uses independent and shared decoders. 2) Using two publicly available pedestrian behavior datasets, namely PIE \cite{Rasouli_2019_ICCV} and JAAD \cite{Rasouli_2017_ICCVW}, we evaluate our model, which achieves state-of-the-art performance on both tasks of trajectory and crossing action prediction. 3) We show the advantages of the components of the proposed model by conducting extensive ablation studies.  

\section{Method}
\subsection{Problem Formulation}
We formulate pedestrian behavior prediction as a multi-objective optimization process in which the goal is to learn distribution $p(L_{p} , a_i, g_i^{t+\tau} | SC_{o}, L_{o}, G_{o}, V)$ for some pedestrian $ 1 < i < n$ where $L_p = \{l^{t+1}_i, l^{t+2}_i, ..., l^{t+\tau}_i\}$, $ a_i \in \{0,1\}$, and $g_i^{t+\tau} \in \{0,1,...,k\}$ are future trajectory, crossing action, and the final location of pedestrian $i$ in a grid on the image plane. Predictions are based on observed scenes  $SC_o = \{sc^{t-m+1}, sc^{t-m + 2}, ..., sc^{t}\}$, the pedestrian's trajectory $L_o =\{l_i^{t-m+1}, l_i^{t-m + 2}, ..., l_i^{t}\}$, their grid locations $ G_o = \{g_i^{t-m+1}, g_i^{t-m + 2}, ..., g_i^{t}\}$ and the ego-vehicle motion $ V = \{v^{t-m+1}, v^{t-m+2}, ..., v^{t+\tau}\}$. Here, $m$ denotes observation duration and $\tau$ is prediction duration.
\subsection{Architecture}
\begin{figure*}
\centering
\includegraphics[width=0.8\textwidth]{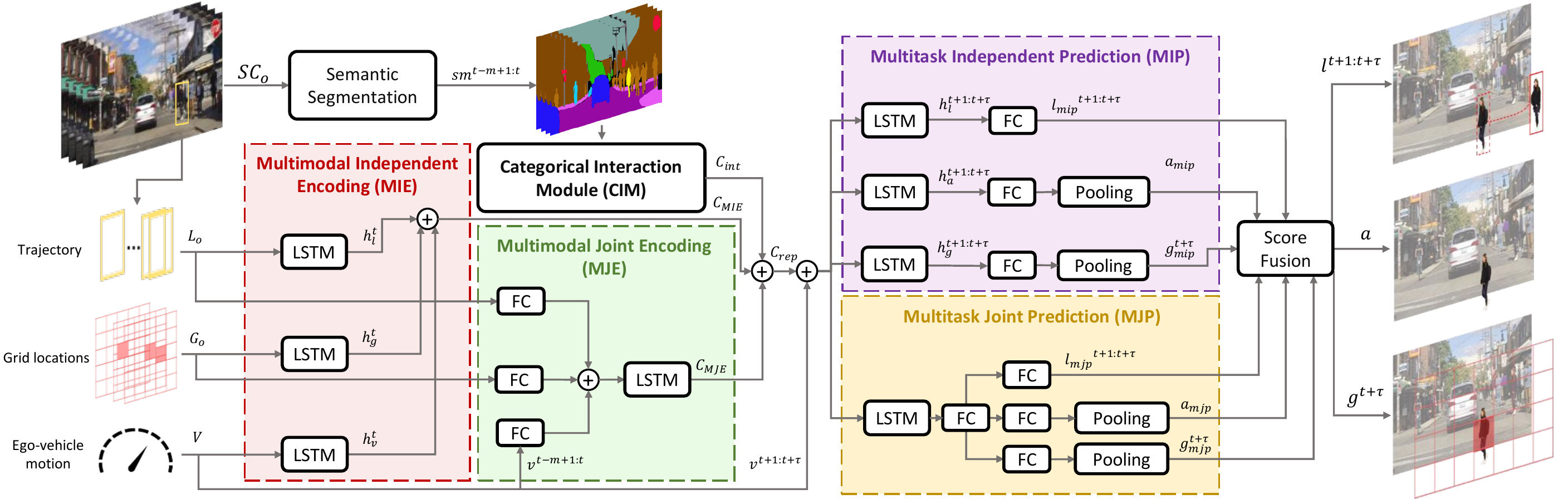}
\caption{The diagram of the proposed multitask pedestrian behavior prediction method. Our model relies on 4 input modalities, namely scene images, and dynamics, including pedestrians' trajectories, their grid locations, and the ego-vehicle motion. The dynamics are encoded both by Multimodal Independent Encoding (MIE) and Multimodal Joint Encoding (MJE) modules. The scene images are converted to semantic maps and fed into Categorical Interaction Module (CIM) to generate interaction representation $C_{int}$. The dynamics encodings are concatenated with $C_{int}$ to form context representation $C_{rep}$, which is combined with planned ego-motion and fed into prediction modules, Multitask Independent Prediction (MIP) and Multitask Joint Prediction (MJP), the outputs of which are averaged for final predictions of pedestrian trajectory ($l^{t+1:t+\tau}$), action ($a$), and final grid location ($g^{t+\tau}$).}
\vspace{-1.5em}
\label{main_diagram}
\end{figure*}

Our approach simultaneously predicts future trajectories, actions and final locations of pedestrians (see Figure \ref{main_diagram}).  Below we discuss different modules of the proposed model:\\
\textbf{Context encoding} deals with processing and encoding of multimodal observation inputs.\\
\textbf{Categorical interaction module} models the interactions between target pedestrians and surrounding traffic elements.\\
\textbf{Behavior prediction} outputs future trajectories of pedestrians, the probabilities of their crossing actions, and final locations in the image grid map.

\subsection{Context Encoding}
We use four different input modalities: scene images, pedestrians' trajectories, their grid locations, and ego-vehicle motion, in order to encode context.

\noindent\textbf{Scenes} are RGB images of traffic scenes recorded from an ego-centric perspective capturing the view in front of the ego-vehicle. These images capture visible changes in the scene. The semantic segmentation module processes scene images and generates semantic maps of the traffic elements $sm^{t-m+1:t}$. These maps are fed into Categorical Interaction Module (CIM) (see Section \ref{cim}) to generate a representation $C_{int}$ that encodes the interactions between target pedestrians and traffic elements.

\noindent\textbf{Pedestrian trajectory.} In the context of prediction in the 2D image plane, in addition to estimating future trajectories, it is also important to localize pedestrian boundaries. As suggested in the past works \cite{Bhattacharyya_2018_CVPR,Rasouli_2019_ICCV}, predicting bounding box coordinates, as opposed to center coordinates, can improve trajectory prediction because bounding box coordinates implicitly capture the changes in relative distance between the ego-vehicle and pedestrians as well as the camera ego-motion. Hence, we use spatial coordinates of bounding boxes around pedestrians defined by top-left and bottom-right corner points $[(x_1,y_1), (x_2,y_2)]$ to encode the changes in the locations of the pedestrians.

\noindent\textbf{Grid locations.} Inspired by \cite{Liang_2019_CVPR}, we convert the location information of pedestrians into grid classes. To achieve this, the image plane is divided into $N \times M$ grid cells each of which is assigned with a unique class. We identify the corresponding grid of pedestrian $i$ at time $t$ by $g^t = \mathrm{argmin}_{j \in cls}(|center_{g_j} - center_{l^t_i}|)$ or the cell whose center is closest to the center of the pedestrian's bounding box. Here, $cls$ refers to classes associated to grid locations.

\noindent\textbf{Ego-vehicle motion} reflects the changes in the state of the ego-vehicle over time denoted as $v^t = [s^t, v_x^t, v_z^t]$ where $s$ is the speed of the vehicle and $v_x$ and $v_z$ are velocities of the vehicle along $x$ and $z$ axes.

\subsubsection{Bifold context representation} In  behavior prediction domain, it is a common practice to process different modalities separately in a unimodal setting, meaning that, first, a separate feature representation is generated for each modality and then these representations are fused prior to inference \cite{Bhattacharyya_2018_CVPR, Liang_2019_CVPR, Rasouli_2019_BMVC}. For example, when using recurrent networks, the last hidden layers, $h^t$, of networks are concatenated. This approach allows each modality to be learned with its own parameters without the noise introduced by other modalities. Independent processing, however, does not capture cross-modal correlations in the temporal dimension and is also potentially susceptible to missing data or noise \cite{Ruder_2017_arxiv}. An alternative is hard parameter sharing, where all modalities are jointly learned using a single model. Parameter sharing can act as a regularizer in a multitask learning framework inducing the model to learn more representative features. To benefit from both approaches, we employ a bifold mechanism to encode input data, namely trajectories, grid locations, and vehicle states. 

\noindent\textbf{Multimodal Independent Encoding (MIE).} This module generates an independent representation for each modality. Each data input is fed into a Recurrent Neural Network (RNN). The last hidden states of the RNNs are concatenated to form a unified representation, $C_{MIE} = h^t_l \oplus h^t_g \oplus h^t_v$, where $\oplus$ is the concatenation operation and $l$, $g$, and $v$ stand for location, grid and vehicle respectively. 

\noindent\textbf{Multimodal Joint Encoding (MJE).} This module jointly encodes different data modalities. Here, it is necessary to project the data from different modalities into a common feature space. MJE generates $ C_{MJE}$ by applying an embedding layer to each input modality, and then concatenating the outputs of embedding layers and processing the concatenated representation using a single RNN.

The final context representation is generated by concatenating all three contextual representations as,
\vspace{-0.15cm}
\begin{equation}
C_{rep} = C_{int} \oplus C_{MIE} \oplus C_{MJE}
\end{equation}

\subsection{Categorical Interaction Module (CIM)}
\label{cim}

\begin{figure}
\includegraphics[width=1\columnwidth]{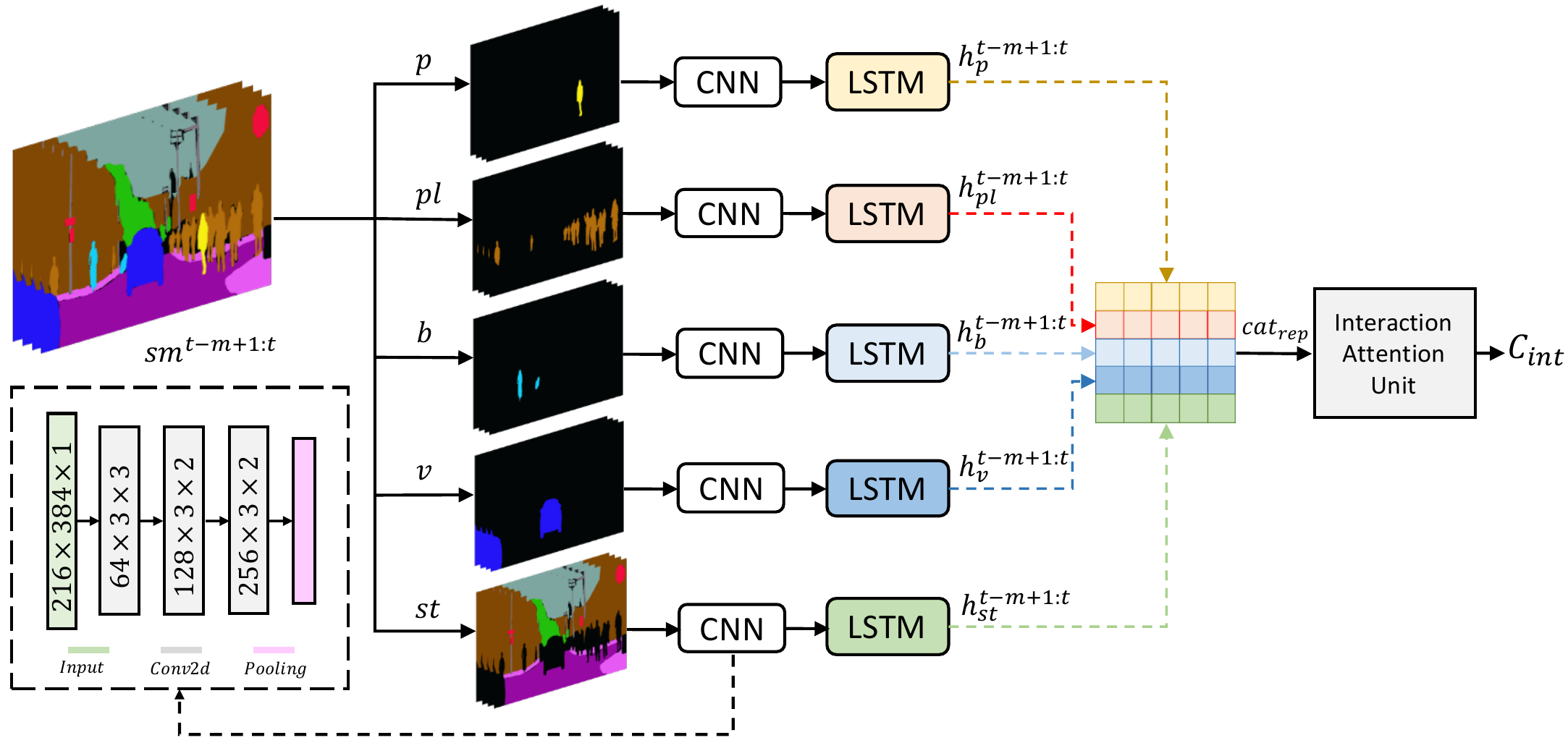}
\caption{Categorical interaction module. Semantic maps are divided into the target pedestrian ($p$), people surrounding the pedestrian ($pl$), motorcyclists/bicyclists ($b$), vehicles ($v$) and static context ($st$). The maps in each category are processed using convolutional layers followed by an LSTM to generate spatio-temporal representations, which are concatenated and fed into the Interaction Attention Unit (IAU) to produce a weighted representation.}
\vspace{-1.5em}
\label{cim_diagram}
\end{figure}

As discussed earlier, in an ego-centric setting without depth information the trajectories of pedestrians in the image plane are not sufficient, and perhaps misleading, for modeling interactions between different agents. To remedy this issue, we rely on semantic parsing of the scenes and implicitly model interactions between target pedestrians and different groups of traffic elements (see Figure \ref{cim_diagram}). We first generate semantic maps using the input scene images to identify the position and category of each object. Then, the maps are divided into different categories, namely the target pedestrian ($p$),  people surrounding the pedestrian ($pl$), motorcyclists/bicyclists ($b$), vehicles ($v$) (e.g. cars, buses, trucks) and static context ($st$) (e.g. signs, roads, signals).

Semantic categories are processed using multiple 2D convolution layers followed by recurrent networks. The hidden states of the RNNs are concatenated to form a shared categorical representation as follow: 
\vspace{-0.1cm}
\begin{equation}
cat_{rep} = h^{o}_p \oplus h^{o}_{pl} \oplus h^{o}_b \oplus h^{o}_v \oplus h^{o}_{st} \in \mathbb{R}^{m \times f}
\end{equation}
\vspace{-0.2cm}

\noindent where observation length $o$ is $m$ and $f$ is the total size of hidden units in recurrent networks. The shared representation is fed into Interaction Attention Unit (IAU) to generate categorical interaction context $C_{int}$.

\noindent\textbf{Interaction Attention Unit (IAU).} Inspired by \cite{Luong_2015_arxiv}, IAU is an attention method that receives as input temporal data and outputs a unified weighted representation. Denoting $h^t$ as the last time step of the input sequence to the attention unit, we first generate attention scores by measuring the similarity between the last and every other time steps,
\begin{equation}
 s^i = {h^t}' W_a h^i
\end{equation}
\noindent where $'$ is the transpose operation. Using the scores, attention weights per time step are  computed by  $\alpha^i = \mathrm{softmax}(s^i)$ and used to calculate the context vector as
\begin{equation}
 c^t = \sum_{i \in [t-m+1:t]}\alpha^{i}h^i.
\end{equation}
\vspace{-0.3cm}

A combination of the context vector and last time step representation is used to generate interaction context,
\vspace{-0.15cm}
\begin{equation}
 C_{int} = \mathrm{tanh}(W_c[c^t\oplus h^t])
\end{equation}
where $C_{int} \in \mathbb{R}^ {1\times q}$ and $\oplus$ is the concatenation operation. 

\subsection{Behavior Prediction}
\label{beh_prediction}
Predictions are made based on the  concatenation of context representation, $C_{rep}$ with future ego-vehicle motion $v^{t+1:t+\tau}$. As in the encoding step, we use a bifold mechanism with two modules: Multitask Independent Prediction (MIP) and Multitask Joint Prediction (MJP).

\noindent\textbf{Multitask Independent Prediction (MIP).} A separate recurrent decoder branch is used for each task to produce three predictions: $l_{mip}^{t+1:t+\tau}$, $a_{mip}$ and $g^{t+\tau}_{mip}$ that represent future trajectories, action and final location of the pedestrian on the grid respectively. 

\textit{Trajectories} are 2D bounding boxes defined by $[(x_1,y_1),(x_2, y_2)]$ denoting the top-left and bottom-right corners of each box. Predictions are made by a linear transformation of hidden states of the trajectory branch.

\textit{Action.} Since the focus of this paper is on intelligent driving systems, we predict pedestrian crossing action, i.e. at a given time, we predict whether the pedestrian will cross in front of the ego-vehicle. To estimate the probability of pedestrian crossing at each time step we perform a linear transformation followed by a sigmoid activation. Then, a global average pooling is used to calculate the mean of predictions over all time steps as the final prediction probability. More formally, the future action is given by,
\vspace{-0.15cm}
\begin{equation}
a_{mip} = \frac{\sum_{i \in [t+1:t+\tau]} \sigma (f(h^i))}{\tau}.
\end{equation}
\vspace{-0.3cm}

\textit{Final grid location} prediction is an auxiliary task. As argued in \cite{Liang_2019_CVPR}, grid location can act as a bridge between trajectory and action prediction and perform as a regularizer by indicating a final destination for the predicted trajectory and the possibility of an action, in this case crossing, i.e. whether a pedestrian's final location falls on the road in the path of the ego-vehicle. Unlike the previous approach, for computational efficiency, we only use a single scale grid set based on the range of pedestrians' scales in the image plane (see Section \ref{ablation_study} for an ablation study).  We treat the grid prediction task as a classification problem and predict the class of the final grid location on the image plane. Following the same procedure as action prediction task, the final grid location at each time step is given by $\mathrm{softmax}(f(h^t))$ and averaged over all time steps.

\noindent\textbf{Multitask Joint Prediction (MJP).} Unlike MIP, this module uses a single RNN as a shared decoder, the output of which is processed using a fully connected (\textit{fc}) layer followed by three separate branches for each task. The predictions are made the same way as in MIP. 

\noindent\textbf{Score fusion.} In order to calculate the final prediction scores for each task, we follow the approach in two-stream methods as in \cite{Carreira_2017_CVPR}, and compute the final scores as follows: $ l^k = \frac{1}{N}\sum_i l^k_i$ where $k = t+1,...,t+\tau$ and $f= \frac{1}{N}\sum_if_i$, where $f \in \{a,g^{t+\tau}\}$, $N =2$ and $i\in\{mip,mjp\}$.


\subsection{Learning Objectives}
The model is trained end-to-end using a multi-objective loss function. For trajectory prediction we use, 
\vspace{-0.15cm}
\begin{equation}
L_{l} = \sum_{i=1}^n \sum_{j=t}^{t+\tau} \log(\mathrm{cosh}(y^j_i - \hat{y}^j_i)),
\end{equation}
\vspace{-0.3cm}

\noindent which compared to commonly used L2 loss as in \cite{Liang_2019_CVPR, Rasouli_2019_ICCV}, is less prone to outliers and generally converges faster. For action prediction, we use a binary cross-entropy loss,
\vspace{-0.15cm}
\begin{equation}
L_{a} = - \sum_{i=1}^n y_i \log(\hat{y}_i) + (1-y_i)\log(1-\hat{y}_i)
\end{equation}
\vspace{-0.3cm}

\noindent and for grid location prediction, use a multi-class entropy, 
\vspace{-0.15cm}
\begin{equation}
L_{g} = - \sum_{i=1}^i \sum_{gc} y_{i,gc} \log(\hat{y}_{i,gc})
\end{equation}
\vspace{-0.3cm}

\noindent where $n$ is the number of samples and $gc$ is the number of grid classes. The final loss is given by,
\vspace{-0.15cm}
\begin{equation}
 L = \alpha L_{l} + \beta L_{a} + \gamma L_g
\end{equation}
\vspace{-0.3cm} 

\noindent where $\alpha$ , $\beta$ and $\gamma$ are loss weights determined empirically.

\section{Evaluation}
\subsection{Implementation}
We use LSTMs for all encoders and decoders with hidden size of 256, L2 regularization of $0.0001$ and $\mathrm{softsign}$ activations with the exception of trajectory decoder for which $\mathrm{tanh}$ is used. The sizes of the embedding layers in MJE and MJP are set to 64 and 128 respectively. For grid classes, the image is divided into a grid of $18 \times 32$ cells of size $60 \times 60$ pixels  based on the lower bound of pedestrian scales in the image plane within the observation horizon.

For semantic maps, we used the method in \cite{Chen_2017_arxiv} pre-trained on the CityScapes dataset \cite{Cordts_2016_CVPR}. The maps are downsampled by a factor of 5 to $384 \times 216$ pixels while maintaining the aspect ratio. We use three 2D convolutional layers (see Figure \ref{cim_diagram} for details) with shared weights across different categories. To encode categories, LSTMs with 128 cells with $\mathrm{tanh}$ activation and $L2$ regularization of $0.0001$ are used. The output dimension of the IAU is set to 128.

\subsection{Datasets}
As the main dataset for our evaluations, we use the \textbf{Pedestrian Intention Estimation (PIE)} dataset \cite{Rasouli_2019_ICCV} which consists of 6 hours of driving footage in urban environments. The dataset provides bounding box annotations for pedestrians and traffic objects as well as sensor readings of the ego-vehicle and ego-motion data recorded from the camera. We use the default data split ratios as in \cite{Rasouli_2019_ICCV}. 

To accommodate both trajectory and action prediction tasks, we clip the pedestrian tracks up to the crossing event frames and sample sequences with $50\%$ overlap and time to event between 1 to 2 seconds (30 to 60 frames) as discussed in \cite{Rasouli_2019_BMVC}. Overall, there are 3980 training sequences out of which 995 are crossing and the rest are non-crossing events. 

We also report on the \textbf{Joint Attention in Autonomous Driving (JAAD)} dataset \cite{Rasouli_2017_ICCVW}, which contains short clips of urban driving scenes. Compared to PIE, the JAAD dataset is less diverse, has shorter sequences and fewer crossing samples,  and does not contain ego-motion data. Instead, JAAD has high-level driver actions, e.g. \textit{moving slow}, \textit{speeding up}, describing the state of the ego-vehicle. We use this information in place of ego-motion data and split the data similar to \cite{Rasouli_2018_ECCVW}. Training samples are generated similar to PIE resulting in 3955 sequences, of which 807 are crossing events.

\subsection{Training}
For training, we use RMSProp \cite{Tieleman_2012_tech} optimizer with initial learning rate of $10^{-4}$ for PIE and $5\times 10^{-5}$ for JAAD and  batch size of 8. We trained the model for 300 epochs and reduced the learning rate by a factor of $0.2$ based on the performance on the validation set. We empirically set $\alpha$, $\beta$ and $\gamma$ values to $0.6$, $1$ and $1$ respectively. To deal with class imbalance for action prediction, we applied class weights based on the ratio of positive and negative samples.  

\subsection{Metrics}
The results are reported for the two primary tasks: trajectory and action prediction with $0.5$s observation length.

\textit{Trajectory prediction}. Following \cite{Malla_2020_CVPR,Liang_2019_CVPR,Gupta_2018_CVPR}, we use two common metrics: Average Displacement Error $\mathrm{ADE}= \frac{\sum_{i=1}^N \sum_{j=t+1}^\tau ||y_i^j - \hat{y}_i^j||_2}{N\times \tau}$
and Final Displacement Error $\mathrm{FDE}= \frac{\sum_{i=1}^N \|y_i^{t+\tau} - \hat{y}_i^{t+\tau}||_2}{N}$. $\mathrm{ADE}$ and $\mathrm{FDE}$ metrics are measured based on the center coordinates of the bounding boxes $[x_c,y_c]$. In addition, to measure the accuracy of bounding box predictions, we report average and final RMSE of bounding box coordinates and denote them as $\mathrm{ARB}$ and $\mathrm{FRB}$ respectively. All metrics are reported in pixels for $1$s prediction length.

\textit{Action prediction}. As in \cite{Rasouli_2019_BMVC, Kotseruba_2021_WACV}, we use common binary classification metrics, namely $\mathrm{accuracy}$, Area Under Curve ($\mathrm{AUC}$), $\mathrm{F1}$ and $\mathrm{precision}$. 

\subsection{Models} 
\noindent \textbf{Trajectory Prediction.} Some past ego-centric trajectory prediction algorithms \cite{Malla_2020_CVPR,Bhattacharyya_2018_CVPR} are compared to well-known methods such as \cite{Gupta_2018_CVPR,Alahi_2016_CVPR}, which are designed and tested on surveillance sequences that are different as they provide bird's eye view of scenes and are recorded using fixed cameras. As a result, we select methods that are trained and tested in a similar ego-centric setting as the proposed algorithms. These methods are \textbf{Future Person Localization (FPL)} \cite{Yagi_2018_CVPR}, 
\textbf{Bayesian LSTM (B-LSTM)} \cite{Bhattacharyya_2018_CVPR}, \textbf{FOL} \cite{Yao_2019_ICRA}, and two variations of the method introduced in \cite{Rasouli_2019_ICCV}, \textbf{PIE$_{traj}$} which only uses bounding boxes for prediction and \textbf{PIE$_{full}$} which is the complete multimodal model. FPL model predicts center coordinates of the bounding boxes, therefore we only report its results on $\mathrm{ADE}$/$\mathrm{FDE}$ metrics.
\noindent\textbf{Action Prediction.} For action prediction, we report the results on state-of-the-art pedestrian crossing prediction algorithms, namely \textbf{ATGC} \cite{Rasouli_2017_ICCVW}, \textbf{MM-LSTM} \cite{Aliakbarian_2018_ACCV}, \textbf{SF-GRU} \cite{Rasouli_2019_BMVC}, and \textbf{PCPA} \cite{Kotseruba_2021_WACV}, for which we pad the context sequence for compatibility with evaluation criteria. Given the similarity between action prediction and recognition tasks, we also use state-of-the-art action recognition model, \textbf{I3D} \cite{Carreira_2017_CVPR}.

\noindent \textbf{Data processing}. A subset of algorithms mentioned above use optical flow and pose information. We use FlowNet 2.0 \cite{Ilg_2017_CVPR} pretrained on \cite{Mayer_2016_CVPR} for optical flow maps and OpenPose \cite{Cao_2017_CVPR} pretrained on \cite{Lin_2014_ECCV} for poses. All these features are generated offline.

\subsection{PIE Dataset}
\subsubsection{Multitask Prediction}
\label{pie_experiment}
\begin{table}[!t]
\vspace{+0.1cm}
\caption{Performance of the proposed method on the PIE dataset. $\uparrow$ and $\downarrow$ mean higher or lower values are better respectively.}

\label{pie_results}
\centering
\resizebox{1\columnwidth}{!}{%
\begin{tabular}{ll|cccc|cccc}
                             &    Method      & $ADE\downarrow$  &$FDE\downarrow$  & $ARB\downarrow$ & $FRB\downarrow$ & $Acc\uparrow$ & $AUC\uparrow$ & $F1\uparrow$ & $Prec\uparrow$ \\ \hline
\multicolumn{2}{l|}{FOL \cite{Yao_2019_ICRA}}      &  73.87   &  164.53 & 78.16   & 143.69 & - & - &  -  &   -    \\
\multicolumn{2}{l|}{FPL \cite{Yagi_2018_CVPR}}       & 56.66    & 132.23  &  -    &   -  & - & - &  -  &   -     \\
\multicolumn{2}{l|}{ B-LSTM \cite{Bhattacharyya_2018_CVPR}}    & 27.09 & 66.74  &  37.41 & 75.87 & - & - &  -  &  -      \\
\multicolumn{2}{l|}{ PIE$_{traj}$ \cite{Rasouli_2019_ICCV}} &21.82  & 53.63 & 27.16& 55.39  & - & - & - & -  \\
\multicolumn{2}{l|}{ PIE$_{full}$ \cite{Rasouli_2019_ICCV}} &19.50  & 45.27 & 24.40& 49.09 & - & - & - & -\\ \hline
\multicolumn{2}{l|}{ ATGC \cite{Rasouli_2017_ICCVW}}&   -   &  -    &  -   &  -  &   0.59  & 0.55    & 0.36   &  0.35      \\
\multicolumn{2}{l|}{ I3D \cite{Carreira_2017_CVPR}}&    -  &   -   &   -  &  -  &0.79&0.75&0.64&0.61    \\
\multicolumn{2}{l|}{ MM-LSTM \cite{Aliakbarian_2018_ACCV}}&   -   &  -    &   -  &  -  &0.84&0.84&0.75&0.68    \\
\multicolumn{2}{l|}{ SF-GRU \cite{Rasouli_2019_BMVC} } &   -   &   -   &   -  &  - &0.86&0.83&0.75&0.73    \\
\multicolumn{2}{l|}{ PCPA \cite{Kotseruba_2021_WACV}} &-&   -   &   -  &  - &0.86&	0.84&	0.76 &	0.73   \\
 \noalign{\hrule height 2pt}
\multicolumn{1}{c|}{\multirow{3}{*}{\rotatebox{90}{Ours}}}&\textbf{BiPed}    & \textbf{15.21}     & \textbf{35.03}  &  \textbf{19.62}   &   \textbf{39.12}  &    \textbf{0.91}  &  \textbf{0.90}  &  \textbf{0.85}   &  \textbf{0.82} \\ 
\multicolumn{1}{l|}{} & BiPed+NEP  & 18.03     & 43.01  &  23.26   &   46.96  &    0.90 &  0.89  &  0.83   &  0.78 \\
\multicolumn{1}{l|}{} & BiPed-NFE  &  18.44    & 45.07  &  24.81 & 50.64    & 0.90    & 0.90   & 0.84   & 0.80  \\

\end{tabular}%

}
\vspace{-0.4cm}
\end{table}

We follow the same evaluation protocol as in \cite{Liang_2019_CVPR} and report the results for single models. For the proposed method, \textbf{Bifold Pedestrian (BiPed)}  prediction, we report on the final model as well as variations of it with no future ego-motion information (\textit{NFE}), and with a noisy ego-motion planner (\textit{NEP}) for which a recurrent decoder, an LSTM similar to other decoders, is used to predict future ego-vehicle motion. 

As illustrated in Table \ref{pie_results}, our method, BiPed, achieves state-of-the-art performance on all metrics. For trajectory prediction, our method significantly improves the results compared to prior state-of-the-art PIE$_{full}$ by up to $22\%$ on $\mathrm{ADE}$ and $20\%$ on $\mathrm{ARB}$ metrics. Significant improvements are also achieved on action prediction. Compared to PCPA, our model achieves $5\%$ and $6\%$ improvements on $\mathrm{Acc}$ and $\mathrm{AUC}$ and even more improvements on $\mathrm{F1}$ and $\mathrm{Prec}$  by $9\%$. These results indicate that the proposed method has more balanced performance compared to others. As expected, when relying on noisy motion predictions, the performance of our model declines. However, it still outperforms PIE$_{full}$ by $8\%$ and $4\%$ on $\mathrm{ADE}$ and $\mathrm{ARB}$ and PCPA by $4\%$ and $7\%$ on $\mathrm{Acc}$ and $\mathrm{F1}$ respectively. It should be noted that our method still achieves state-of-the-art performance on most metrics even without future ego-motion information. 

\begin{figure}
\includegraphics[width=1\columnwidth]{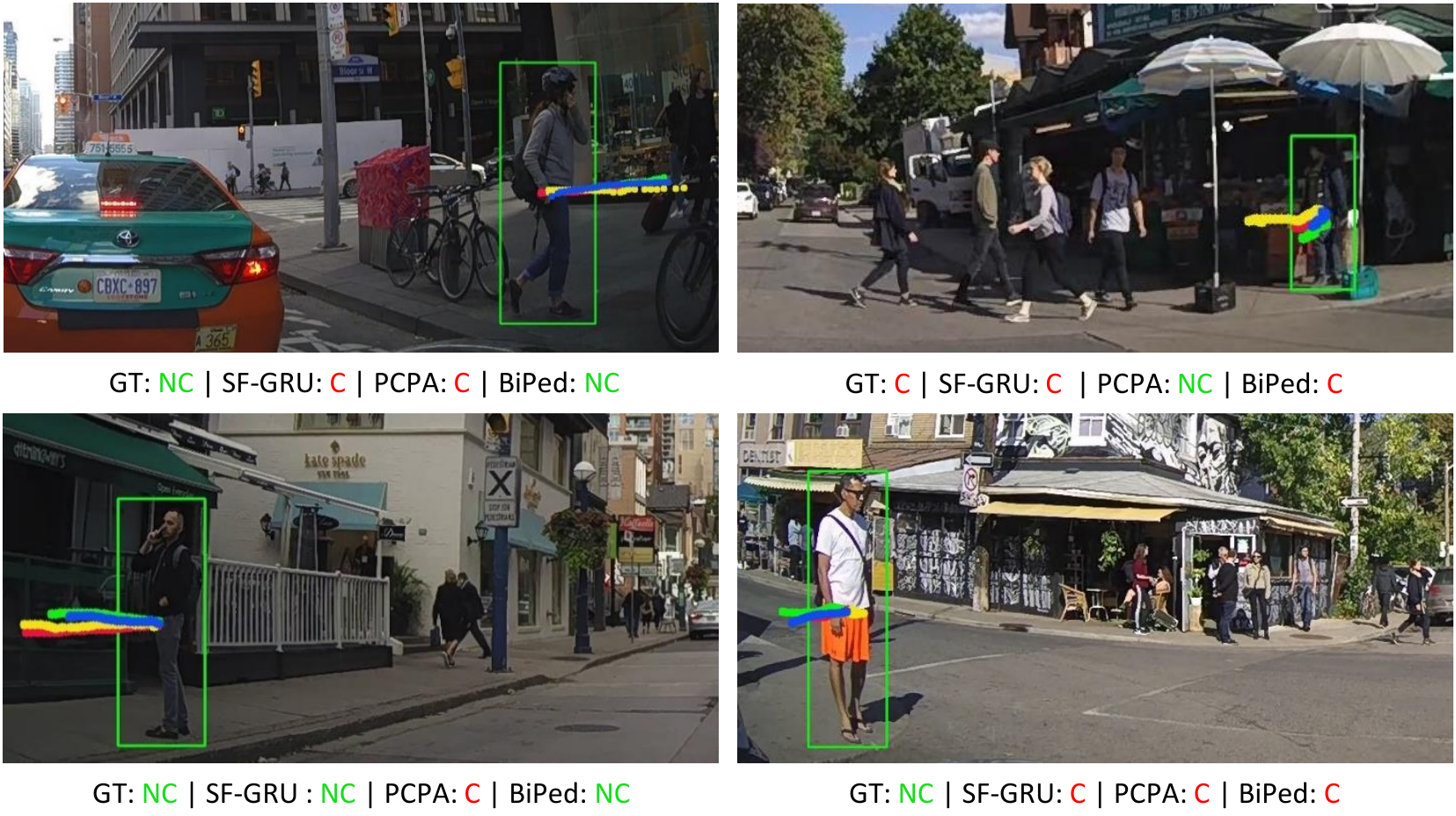}
\caption{Qualitative results of the proposed algorithm on PIE. Trajectories show one second in future and correspond to {\color{gt}ground truth}, {\color{ours} BiPed (\textbf{ours})}, {\color{pie}PIE$_{full}$}, and {\color{b-lstm}B-LSTM}. For actions, we report the results on BiPed (\textbf{ours}), SF-GRU and PCPA. The results correspond to pedestrian crossing ({\color{red}C}) and not-crossing ({\color{green}NC}) actions. Here, ground truth is denoted as GT.}
\label{qualitative}
\vspace{-0.1cm}
\end{figure}

Our method is also more stable compared to the past arts.  For instance, standard deviation of the method over 20 runs with random initialization on $\mathrm{ADE}$ is $0.21$ compared to $0.46$ for PIE$_{full}$ and on $\mathrm{F1}$ is $0.006$ whereas for PCPA is $0.01$. Qualitative examples are illustrated in Figure \ref{qualitative}.

\vspace{-0.2cm}
\subsubsection{Ablation study}
\label{ablation_study}
\textbf{Independent and joint processing.} We examine the impact of different proposed encoding and decoding modules, namely Multimodal Independent Encoding (MIE), Multimodal Joint Encoding (MJE), Multitask Independent Prediction (MIP), and Multitask Joint Prediction (MJP). 
\begin{table}[!t]
\caption{The impact of different encoding and decoding schemes. $\uparrow$ and $\downarrow$ mean higher or lower values are better respectively. }
\vspace{+0.1cm}

\label{ablation_sharing}
\resizebox{\columnwidth}{!}{%
\begin{tabular}{l|cccc|cccc}
            Modules                           & $ADE\downarrow$  &$FDE\downarrow$  & $ARB\downarrow$ & $FRB\downarrow$ & $Acc\uparrow$ & $AUC\uparrow$ & $F1\uparrow$ & $Prec\uparrow$\\ \hline
MIE+MIP           &  15.87   &  35.26   & 20.61  &  39.53  & 0.85   & 0.86    & 0.76   & 0.67   \\
MIE+MJP		 	 &16.22	&36.41&	21.71&	42.22  & 0.89	&0.89	&0.82	&0.77 \\
MJE+MIP          &15.73&35.40&	21.29	&41.07  & 0.86&	0.87	&0.78	&0.70 \\
MJE+MJP         	&	16.53	&36.36	&22.51	&43.47  & 0.87	&0.88&	0.80	&0.72\\
MIE+MJP+MIP     	&15.59	&35.68&	20.52&	40.86   &    0.90 &0.89	&0.83&	0.80 \\ \hline
\textbf{All}   & \textbf{15.21}     & \textbf{35.03}  &  \textbf{19.62}   &   \textbf{39.12} &   \textbf{0.91}  &  \textbf{0.90}  &  \textbf{0.85}   &  \textbf{0.82} \\ 
\end{tabular}%
}
\vspace{-0.5cm}
\end{table}

As shown in Table \ref{ablation_sharing}, MIP and MJP play complementary roles, i.e. MIP results in better trajectory predictions while MJP improves the results on action prediction. Even though trajectories and actions are correlated, the manner in which they are learned together is important. Multiple trajectories can correspond to the crossing action as long as pedestrian and vehicle paths intersect. However, inferring trajectories from the crossing action is more ambiguous because information, such as direction or speed of the pedestrian, is not directly implied by the action. Such ambiguity can increase uncertainty of predicting trajectories when a joint prediction module is used. That is why the best performance on all metrics is achieved when both MIP and MJP methods are combined. In case of encoding modules, MIE and MJE, although when used individually the performance on different tasks do not vary much, when combined they tend to complement each other and boost performance on all metrics as shown in the last two rows of the table.

\noindent\textbf{Interaction encoding.} We examine the contribution of Categorical Interaction Module (CIM) to the overall performance of our method. Here, we only report on trajectory metrics since the variation on action prediction results were insignificant. We consider four alternative feature representation schemes where all classes are either represented in a \textit{Single} semantic map or separated into \textit{Categorical} maps and are processed using only 2D (\textit{Conv2D}) or 3D (\textit{Conv3D}) convolutional layers as specified in Figure \ref{cim_diagram}. For Conv2D versions, we follow the method in \cite{Liang_2019_CVPR} and average the maps along temporal dimension. We refer to our approach which uses both Conv2D layers and LSTMs as \textit{Hybrid}.

\begin{figure}
\includegraphics[width=1\columnwidth]{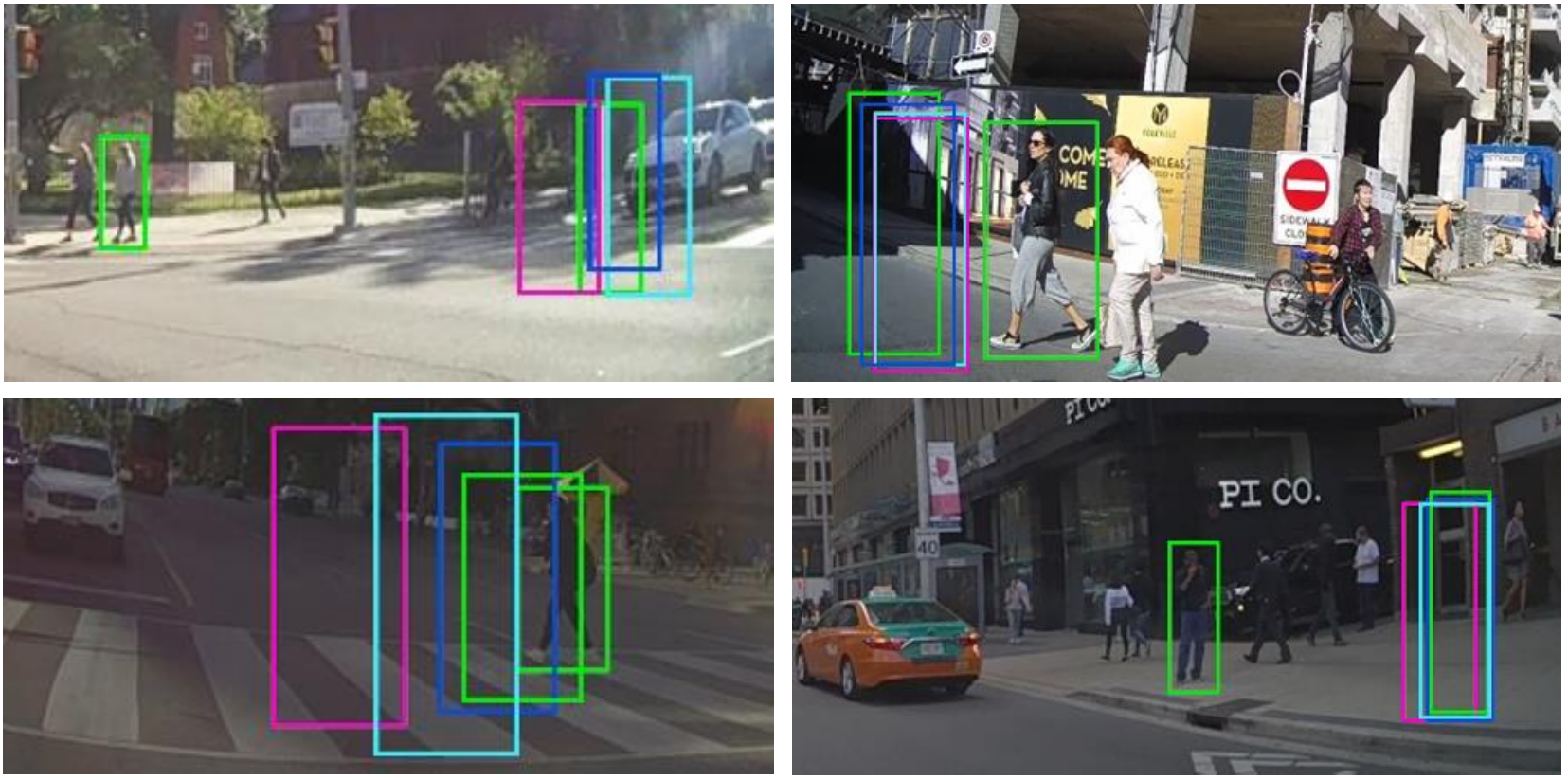}
\caption{Qualitative results of the proposed algorithm on PIE showing the final predicted bounding boxes using {\color{no_cim}no CIM}, {\color{hybrid}Hybrid} and {\color{ours}Hybrid+IAU} CIM modules against {\color{gt}ground truth}.}
\label{cim_qualitative}
\vspace{-0.3cm}
\end{figure}

The results are summarized in Table \ref{ablation_CIM}. Here, we can see that using CIM improves the results by up to $5\%$ on $\mathrm{ADE}$/$\mathrm{FDE}$ and $12\%$ on $\mathrm{ARB}$/$\mathrm{FRB}$. The results also show the advantage of separating semantic maps into categories with shared characteristics. This can be due to the heterogeneous nature of pedestrians' interactions with their surroundings which necessitates learning a separate representation for each type of interaction. Overall, the proposed hybrid approach using categorical semantic representations clearly stands out on all metrics. Using the Interactive Attention Unit (IAU) further improves the results as the model learns to dynamically focus on different aspects of the interaction in a given context (see Figure \ref{cim_qualitative} for some  examples).

\begin{table}[!t]
\caption{Evaluation of alternative interaction modeling methods. $\downarrow$ means lower values are better.}
\vspace{+0.1cm}
\label{ablation_CIM}
\resizebox{\columnwidth}{!}{%
\begin{tabular}{l|cccc}
      Interaction Model            & $ADE\downarrow$  &$FDE\downarrow$  & $ARB\downarrow$ & $FRB\downarrow$   \\ \hline
No CIM            & 16.00   & 36.89 & 22.11  & 44.46 \\
Single Conv2D     & 17.04 & 39.65 & 23.77  & 48.00  \\
Single Conv3D     & 17.05 & 40.38 & 22.42  & 45.78 \\
Categorical Conv2D            & 15.70  & 36.28 & 21.04 & 41.84 \\
Categorical Conv3D            & 15.85 & 36.95 & 22.00     & 44.53 \\\hline
Ours (Single Hybrid)    & 16.17 & 37.17 & 21.39  & 42.21  \\ 
Ours (Categorical Hybrid)    & 15.55 & 35.96 & 20.36  & 40.60  \\ 
\textbf{Ours (Categorical Hybrid+IAU) }& \textbf{15.21} & \textbf{35.03} & \textbf{19.62}  & \textbf{39.12}
\end{tabular}
}
\vspace{-0.4cm}
\end{table}

\begin{table}[!t]

\caption{Ablation study on grid classification task. Grid cell (GC) sizes are in $px^2$. $\uparrow$ and $\downarrow$ mean higher or lower values are better respectively.}
\vspace{+0.1cm}
\label{grid_results}
\centering
\resizebox{1\columnwidth}{!}{%
\begin{tabular}{l|cccc|cccc}
 Method      & $ADE\downarrow$  &$FDE\downarrow$  & $ARB\downarrow$ & $FRB\downarrow$ & $Acc\uparrow$ & $AUC\uparrow$ & $F1\uparrow$ & $Prec\uparrow$ \\ \hline
 No grid  &   16.38 &	36.60 & 22.10&	 43.95  &  0.89&	0.89	&0.82&	0.76 \\ 
 GC-15  &  16.30   & 35.88  & 21.41  & 42.82    & 0.90   & \textbf{0.90}   & 0.84   & 0.78  \\ 
 GC-30  &  15.71&	36.01&	20.85&	41.35    & \textbf{0.91}   & \textbf{0.90}   &  0.84  & 0.81  \\ 
 \textbf{GC-60}    & \textbf{15.21}     & \textbf{35.03}  &  \textbf{19.62}   &   \textbf{39.12}  &    \textbf{0.91}  &  \textbf{0.90}  &  \textbf{0.85}   &  \textbf{0.82} \\ 
 GC-120  &16.32&	 37.15&	21.38&	42.43   & \textbf{0.91}  &  0.89  & 0.84   & \textbf{0.82}  \\ 

\end{tabular}%

}
\vspace{-0.4cm}
\end{table}

\noindent\textbf{Grid classification task.} We evaluate the proposed model with no auxiliary grid task (\textit{no grid}) and different grid cell (\textit{GC}) sizes  in $px^2$. As shown in Table \ref{grid_results}, overall, grid classification task is beneficial for both trajectory and action predictions, but improvements vary depending on the grid resolution. The best performance is achieved at grid cell size of $60$ on all metrics, where $\mathrm{ADE/ARB}$ is improved by $7\%$ and $\mathrm{precision}$ by $6\%$. When the grid cells are too large, e.g. $120$, they are not effective, particularly on trajectories, which are affected the most because multiple time steps of the pedestrian movement may fall within a single cell. 


\begin{table}[!t]
\caption{Performance of the proposed method on JAAD. $\uparrow$ and $\downarrow$ mean higher or lower values are better respectively. }

\label{jaad_perf}
\resizebox{\columnwidth}{!}{%
\begin{tabular}{ll|cccc|cccc}
 & Method       & $ADE\downarrow$  &$FDE\downarrow$  & $ARB\downarrow$ & $FRB\downarrow$ & $Acc\uparrow$ & $AUC\uparrow$ & $F1\uparrow$ & $Prec\uparrow$\\ \hline
\multicolumn{2}{l|}{ FOL \cite{Yao_2019_ICRA}}    & 61.39&126.97&  70.12& 129.17&  -  &  -  & - & -\\
\multicolumn{2}{l|}{ FPL \cite{Yagi_2018_CVPR}}     & 42.24   &  86.13 &  - & - &  -  &  -  & - & - \\
\multicolumn{2}{l|}{ B-LSTM \cite{Bhattacharyya_2018_CVPR}} & 28.36 & 70.22  &  39.14 & 79.66 &  -  &  -  & - & -   \\
\multicolumn{2}{l|}{PIE$_{traj}$ \cite{Rasouli_2019_ICCV}} & 23.49& 50.18 &30.40  & 57.17 & - & - &- & -  \\
\multicolumn{2}{l|}{PIE$_{full}$ \cite{Rasouli_2019_ICCV}} & 22.83& 49.44	&29.52&	55.43 & - & - &- & -  \\ \hline 
\multicolumn{2}{l|}{ATGC \cite{Rasouli_2017_ICCVW}} & - & - & - & -&0.64 &0.60& 0.53& 0.50    \\
\multicolumn{2}{l|}{I3D \cite{Carreira_2017_CVPR}}  &    -  &   -   &   -  &  -   &  0.82   & 0.75 & 0.55   &  0.49  \\
\multicolumn{2}{l|}{MM-LSTM \cite{Aliakbarian_2018_ACCV}} &   -   &  -    &   -  &  -   &  0.80  & 0.60 & 0.40   &  0.39  \\
\multicolumn{2}{l|}{SF-GRU \cite{Rasouli_2019_BMVC}} &   -   &   -   &   -  &  -   &  \textbf{0.83}  & 0.77 &  0.58  &  0.51 \\ \multicolumn{2}{l|}{ PCPA  \cite{Kotseruba_2021_WACV}} &   -   &   -   &   -  &  -   &  \textbf{0.83}  & 0.77 & 0.57   &  0.50  \\
\noalign{\hrule height 1.5pt}
\multicolumn{1}{l|}{\multirow{3}{*}{\rotatebox{90}{Ours}}} &\textbf{BiPed} &  \textbf{20.58}  & \textbf{46.85}  &  \textbf{27.98}   &  \textbf{55.07} & \textbf{0.83}     & \textbf{0.79}  &  \textbf{0.60}   &   \textbf{0.52}  \\ 
\multicolumn{1}{l|}{} &BiPed+NEP    &    20.75  &  47.44  &  28.16  &  55.50 & 0.83     & 0.79  &  0.60   &   0.51 \\
\multicolumn{1}{l|}{} &BiPed+NFE   &  21.13& 48.88&	29.98&	56.52 &  0.83   & 0.78   & 0.59   & 0.52    
 
\end{tabular}%
}
\vspace{-0.5cm}
\end{table}
\subsection{JAAD Dataset}
We follow the same procedure as in Section \ref{pie_experiment} and evaluate our model on the JAAD dataset. As shown in Table \ref{jaad_perf}, the performance improvement is smaller, particularly on action prediction, due to the fact that compared to PIE, the JAAD dataset is less diverse, less balanced, and does not contain the ego-vehicle motion information. However, for trajectory prediction, our model clearly stands out by improving over PIE$_{full}$ by up to $10\%$ on $\mathrm{ADE}$ and $5\%$ on $\mathrm{ARB}$, while maintaining state-of-the-art performance on action prediction with $2\%$ improvement on $\mathrm{AUC}$ and $\mathrm{F1}$ metrics.  Using a noisy motion planner on JAAD, the performance decline is negligible because predictions are made on a small set of driver's actions as opposed to continuous  velocity of the vehicle provided in PIE.
\vspace{-0.2cm}
\section{Conclusion}
We presented a multitask learning framework for predicting pedestrian trajectory and action. Our method relies on a bifold mechanism to encode and decode different input modalities and tasks, thus allowing the model to learn cross-correlation between them and inducing it to learn better representations. In addition, we introduced a novel technique which implicitly models interactions between target pedestrians and their surroundings by relying on changes in semantic representations of the scenes. Using publicly available benchmarks, we showed that our proposed method significantly improves over existing methods on both trajectory and action prediction tasks. We further showed the overall contributions of our novel modules by conducting ablation studies. The proposed approach can be applied to human behavior understanding in other computer vision and robotics tasks, such as action and gesture recognition, interaction prediction, and group activity understanding.

\bibliographystyle{IEEEtran}
\bibliography{biped}
\end{document}